\documentclass{article}
\usepackage{spconf}
\usepackage{url}
\usepackage[T1]{fontenc}
\usepackage{cite}
\usepackage{amsmath,amssymb,amsfonts}
\usepackage{graphicx}
\usepackage{textcomp}
\usepackage{xcolor}
\usepackage{eso-pic}
\usepackage{booktabs,array}
\usepackage{comment}
\usepackage{float}
\usepackage[hidelinks]{hyperref}
\let\oldthebibliography\thebibliography
\renewcommand{\thebibliography}[1]{%
  \oldthebibliography{#1}%
  \setlength{\itemsep}{0pt}%
  \setlength{\parsep}{0pt}%
}

\newcommand{\bd}[1]{\textbf{#1}}

\newcommand{\condnf}{$\mathrm{C_{95}}$}
\newcommand{\condzero}{$\mathrm{C_0}$}
\newcommand{\condp}{$\mathrm{C_p}$}
\newcommand{\condph}{$\mathrm{C_p^{hard}}$}

\title{ExpandDiff: Dynamic Range Expanding Diffusion for Single-Image HDR Reconstruction}

\name{Mehmet Emre And{\i}ran$^{*}$, Zhuoqian Yang, Liying Lu, Mathieu Salzmann, Sabine S\"usstrunk
}

\address{School of Computer and Communication Sciences, EPFL, Switzerland \\
firstname.lastname@epfl.ch, $^{*}$memreandiran@gmail.com}

\newif\ifarxiv
\arxivtrue

\begin{document}

\ifarxiv
\AddToShipoutPictureFG*{%
  \AtPageLowerLeft{%
    \put(\LenToUnit{0.5\paperwidth},\LenToUnit{0.45in}){%
      \makebox[0pt][c]{\parbox[b]{\textwidth}{\centering\scriptsize
        This work has been submitted to the IEEE for possible publication.
        Copyright may be transferred without notice, after which this
        version may no longer be accessible.}}}}}
\fi
\ninept%
\maketitle

\begin{abstract}
Single-image HDR reconstruction requires inferring missing detail while preserving the visible content of an LDR image. Differences in sensor dynamic range and exposure cause LDR images to lose varying amounts of information in shadows and highlights. We present ExpandDiff, a conditional diffusion pipeline that jointly reconstructs clipped shadows and highlights. To account for this variation, we introduce Dynamic Clipping Synthesis (DCS), which randomly samples shadow and highlight clipping percentiles when constructing training inputs from HDR targets. A pixel-space diffusion model guided by spatially-adaptive normalization then predicts perceptually encoded HDR through a bounded output head, reconstructing both clipping directions in one sampling trajectory. On the SI-HDR benchmark, ExpandDiff variants improve HDR reconstruction accuracy by 3.43\,dB in PU21-PSNR over the strongest evaluated competing method, and by 7.34\,dB under two-sided clipping. The code and supplementary material are available at \url{https://memreandiran.github.io/expanddiff/}.
\end{abstract}


\section{Introduction}
\label{sec:intro}

Typical camera sensors capture 8 to 12 stops of dynamic range~\cite{eilertsen2017}, while real scenes can span 20 stops or more~\cite{fairchild2007}. The sensor's dynamic range determines the span of scene intensities that can be recorded, while exposure determines where this span lies. LDR images captured with different sensors and exposures can therefore lose different amounts of information in shadows, highlights, or both. Single-image HDR reconstruction must accommodate this variation, inferring plausible detail in regions where information is lost while preserving visible image content. This inference is ill-posed because the missing detail cannot be uniquely determined from the LDR image. Earlier methods emphasize saturated highlights~\cite{eilertsen2017,santos2020} or reconstruct HDR through intermediate exposure brackets~\cite{endo2017,lee2018,lediff2025}. We directly predict one HDR image, jointly restoring shadows and highlights under varying clipping levels.

We present \emph{Dynamic Range Expanding Diffusion} (ExpandDiff), a pipeline that couples variable clipping during training, spatial LDR guidance, and perceptually encoded HDR prediction. How much a capture is clipped varies from image to image, and the model is not told it at inference. To account for variation in clipping across real captures, we introduce \emph{Dynamic Clipping Synthesis} (DCS): a training strategy that randomly samples shadow and highlight clipping levels when constructing LDR guidance from HDR targets. DCS exposes a single model to varying amounts of information loss at both ends of the dynamic range, encouraging reconstruction across diverse clipping conditions. We adapt the RGB-guided RAW-Diffusion backbone~\cite{reinders2025rawdiffusion} to predict HDR RGB images. The backbone uses spatially-adaptive normalization (SPADE)~\cite{park2019spade} to inject LDR features into the diffusion network, so our model conditions the diffusion process in pixel space rather than through the conditioning latent of \cite{dalal2023} or the compressed latent space of \cite{lediff2025,Goswami_2025,wu2026x2hdr}.

Finally, to construct the training target, we encode the HDR target with PU21~\cite{mantiuk2021pu21}, a perceptually uniform encoding of absolute luminance in which equal differences in its units are equally visible. This allocates more of the bounded output range to dark intensities and prevents the loss function from getting dominated by highlights. The model retains Gaussian diffusion training and reconstructs shadows and highlights jointly in a single 24-step sampling trajectory.

Our contribution is the integrated dynamic range expansion pipeline, rather than any single component, and its evaluation across clipping distributions. We evaluate on the 181-scene SI-HDR benchmark~\cite{hanji2022}, along with additional degradations applied to the same HDR references. The benchmark-oriented variant, ExpandDiff-B, achieves the best gain-aligned PU21-PSNR among the evaluated methods, exceeding the strongest competing method by a paired mean of $3.43$\,dB. ExpandDiff-P, trained with DCS, ranks second on that metric and exceeds the strongest competing method by $7.34$\,dB under two-sided clipping. 

\section{Related Work}
\label{sec:related}

\textbf{Single-image HDR reconstruction.} Eilertsen et al.~\cite{eilertsen2017} reconstruct saturated highlights with a CNN. ExpandNet~\cite{marnerides2018} combines features at multiple spatial scales, Santos et al.~\cite{santos2020} use feature masking and perceptual supervision, and Liu et al.~\cite{liu2020} learn to reverse the camera pipeline. Other methods synthesize exposure brackets and merge them into HDR~\cite{endo2017,lee2018}. These approaches establish synthetic degradation and recovery of under- and overexposure as training objectives. DCS samples shadow and highlight clipping levels in percentile space, controlling the fraction of affected pixels across scenes. We couple this training distribution with spatially guided diffusion and a perceptual prediction domain.

\noindent \textbf{Diffusion-based inverse tone mapping.} Dalal et al.~\cite{dalal2023} condition diffusion on an encoded latent of the input for LDR-to-HDR conversion, while Goswami et al.~\cite{Goswami_2025} incorporate semantic guidance. LEDiff~\cite{lediff2025} fuses exposure brackets from separate highlight and shadow denoisers in a pretrained LDR latent space. ExpandDiff instead predicts one HDR image in pixel space, reconstructing both clipping directions in the same sampling trajectory. ExpoCM~\cite{liu2026expocm} also handles different exposure regimes jointly, using exposure-aware consistency trajectories for one-step inference.

Further discussion of related works is provided in the supplementary material.

\section{Method}
\label{sec:method}

\begin{figure}[t]
\centering
\includegraphics[width=\columnwidth]{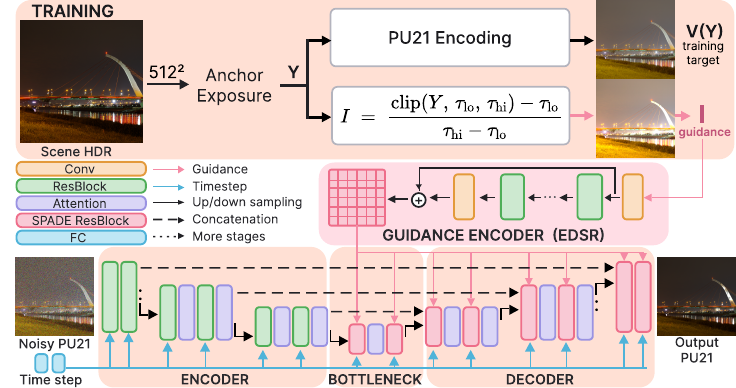}
\caption{ExpandDiff pipeline. Dynamic Clipping Synthesis varies the clipping levels of the LDR guidance during training, while PU21 encoding defines the HDR prediction domain. Spatial guidance conditions the diffusion network, and inverse encoding yields linear HDR.}
\label{fig:method}
\end{figure}

Fig.~\ref{fig:method} summarizes ExpandDiff. We first construct a PU21 encoded HDR target and use Dynamic Clipping Synthesis (DCS) to synthesize its LDR guidance. A conditional diffusion model then learns to recover the perceptually encoded target. At inference, the LDR image guides a single sampling trajectory, and the predicted image is decoded to linear HDR.

\subsection{Dynamic Clipping Synthesis}
\label{sec:traindist}

The clipping level of a real input is unknown at inference, and it differs with the exposure, the scene, and the camera. DCS varies the amount of information removed from shadows and highlights as training examples are constructed. This enables a single checkpoint to perform dynamic range expansion even when it receives images with varying degrees of degradation as input. For each HDR target $Y$, we synthesize the linear guidance image
\begin{equation}
I=\frac{\operatorname{clip}(Y,\tau_{\mathrm{lo}},\tau_{\mathrm{hi}})-\tau_{\mathrm{lo}}}
{\tau_{\mathrm{hi}}-\tau_{\mathrm{lo}}}.
\label{eq:clip}
\end{equation}

For each training example, DCS draws the clipping percentiles $q_{\mathrm{lo}}\sim\mathcal{U}[0,10]$ and $q_{\mathrm{hi}}\sim\mathcal{U}[0,30]$, in percent. $\tau_{\mathrm{lo}}$ is the $q_{\mathrm{lo}}$ percentile of the per-pixel channel minima, and $\tau_{\mathrm{hi}}$ is the $(100-q_{\mathrm{hi}})$ percentile of the per-pixel channel maxima. The same scalar thresholds apply to all channels. The percentiles control the fraction of affected pixels, allowing the training distribution to span near-unclipped through heavily clipped inputs across scenes with different intensity distributions. Neither the thresholds nor clipping masks are supplied to the network, so it must infer the required expansion from the LDR image itself.

For data, we use HDR images from Poly Haven, the public Laval photometric sample~\cite{bolduc2023}, HDR-Real~\cite{liu2020}, and Fairchild~\cite{fairchild2007}. We form $512\times512$ images by center-cropping ordinary images and extracting five random perspective views from each panorama. Each resulting image is rescaled to a median luminance of $20$\,cd/m$^2$ and clipped at a peak of $L_{\max}=1000$\,cd/m$^2$. Dividing by this peak gives a target $Y\in[0,1]^{3\times512\times512}$. This defines a consistent display-referred output scale.

\subsection{PU21 Prediction and the Bounded Head}
\label{sec:pu21}

PU21 is a perceptually uniform encoding of absolute luminance \cite{mantiuk2021pu21}, where equal differences in its units are equally visible. We predict in the PU21 space and use it to supervise our model to exploit this property. PU21 is important for supervision, as in linear radiance, a perceptually equal error is numerically far larger in the highlights than in the shadows, so a pixelwise distance loss would be dominated by bright pixels. On PU21 values, each error instead counts according to its visibility.

Let $U$ denote the PU21 encoding of absolute linear intensity~\cite{mantiuk2021pu21}. We apply it per channel and normalize at the target peak:
\begin{equation}
 V(Y)=\frac{U(L_{\max}Y)}{U(L_{\max})},\qquad x_0=2V(Y)-1.
\label{eq:pu21}
\end{equation}
The network's $\tanh$ head bounds $\hat x_0$ to $[-1,1]$. We decode the final prediction as $\hat Y=V^{-1}((\hat x_0+1)/2)$; the guidance $I$ remains linear.

This encoding changes the operating range of the $\tanh$ output head. With a linear target spanning $0$ to $1000$\,cd/m$^2$, an intensity of $0.5$\,cd/m$^2$ maps to $-0.999$, near the saturated tail of $\tanh$. PU21 assigns more of the output range to dark intensities, placing them where $\tanh$ is more responsive, which improves the head's sensitivity to shadow errors. We test the effects of the encoding and head jointly in Sec.~\ref{sec:ablation}.

\subsection{Spatially Conditioned Diffusion Backbone}
\label{sec:backbone}

HDR reconstruction must preserve visible structure while inferring content in clipped regions, motivating a pixel-space diffusion backbone with spatial LDR guidance at multiple resolutions. We adapt RAW-Diffusion's conditional U-Net~\cite{reinders2025rawdiffusion} from RAW Bayer prediction to HDR RGB prediction. An EDSR encoder~\cite{lim2017edsr}, without its upsampling head, extracts a full-resolution guidance feature map. SPADE layers~\cite{park2019spade} resample these features to each bottleneck and decoder resolution and predict spatial scale ($\gamma$) and shift ($\beta$) maps. For a group-normalized activation $h$, conditioning gives $(1+\gamma(I))\odot h+\beta(I)$. Thus, image structure is available throughout reconstruction without compressing the HDR prediction into a VAE latent space.

The U-Net has six resolution levels, channel widths from $32$ to $128$, two residual blocks per level, and self-attention at the two coarsest resolutions and the bottleneck. The guidance encoder uses four residual blocks with $64$ channels. ExpandDiff has $25.0$\,M parameters. We use 24 DDIM steps~\cite{song2021ddim}; both clipping directions are reconstructed by the same network and trajectory. An 8-bit inference input is linearized with the inverse of the sRGB transfer function, regardless of its original transfer function, and mapped into the range $[0,1]$. The resulting image is approximately linear, even though the actual camera response is unknown.

Let $x_0=2V(Y)-1$ denote the normalized PU21 target, defined in Sec.~\ref{sec:pu21}. We use the standard Gaussian forward process,
\begin{equation}
 x_t=\sqrt{\bar\alpha_t}\,x_0+\sqrt{1-\bar\alpha_t}\,\epsilon,
 \qquad \epsilon\sim\mathcal{N}(0,\mathbf{I}),
\label{eq:diffusion}
\end{equation}
where $\bar\alpha_t$ is the cumulative signal-retention factor. Following RAW-Diffusion~\cite{reinders2025rawdiffusion}, the network predicts the clean target, $\hat x_0=f_\theta(x_t,t,I)$. Its objective is
\begin{equation}
\mathcal{L}=\mathbb{E}\!\left[
\frac{\operatorname{MAE}(\hat x_0,x_0)+\operatorname{MSE}(\hat x_0,x_0)}
{1+\operatorname{SNR}(t)}\right],
\label{eq:loss}
\end{equation}
with $\operatorname{SNR}(t)=\bar\alpha_t/(1-\bar\alpha_t)$. We retain equally weighted absolute and squared errors, but omit RAW-Diffusion's logarithmic error term because our target is already perceptually encoded.

\section{Experiments}
\label{sec:setup}

We compare ExpandDiff with published models on SI-HDR~\cite{hanji2022}, then test transfer across clipping levels and directions. A factorial ablation examines PU21 encoding and the bounded prediction head.

\subsection{Evaluation Setup}
\label{sec:conds}

\textbf{Data and conditions.} SI-HDR provides 181 scenes with $1280\times1888$ HDR references. We center-crop and box-filter them to $512\times512$, the input size used for all methods, as LEDiff~\cite{lediff2025} only accepts this size. We evaluate the benchmark's supplied \texttt{clip\_95} inputs (\condnf{}) after the same crop and resize, in which $8.17\%$ of pixels have a channel at the ceiling and none at zero.

To test reconstruction under degradation at both ends, we construct \condp{} from the same references with Eq.~\eqref{eq:clip} at fixed $q_{\mathrm{lo}}=5$, $q_{\mathrm{hi}}=15$, the midpoints of P's training ranges. On average, $15.40\%$ of pixels are highlight-clipped, and $5.87\%$ are shadow-clipped. Every image contains both types of clipping. Thus, \condnf{} tests highlight reconstruction under the benchmark pipeline, while \condp{} tests joint shadow and highlight reconstruction within the clipping ranges used by DCS. We also evaluate doubled clipping percentiles (\condph{}) and LEDiff's independently defined degradation (\condzero{})~\cite{lediff2025, andersson2021hdrflip}; details appear in the supplementary material. 

\noindent \textbf{Models.} ExpandDiff-P is trained with DCS (Sec.~\ref{sec:traindist}); the P suffix denotes its percentile-based clipping distribution. ExpandDiff-B uses a benchmark-oriented degradation with a highlight-only clip at $q_{\mathrm{hi}}\sim\mathcal{U}[3,7]$, an 8-bit round trip, and a randomized two-parameter response curve. They share data sources, architecture, target encoding, loss, and batch size $32$, but are trained for 150k and 75k steps at learning rates $2\times10^{-4}$ and $10^{-4}$, respectively. We compare with ExpandNet~\cite{marnerides2018}, MaskHDR~\cite{santos2020}, the highlight and shadow variants of LEDiff (HL and SH)~\cite{lediff2025}, and Refusion-HDR and DITM from AIM 2025~\cite{aim2025itm}, giving six configurations across five methods. We run the released models and include the unprocessed LDR input, a baseline many SI-HDR methods fail to improve on~\cite{hanji2022}. All methods receive the same 8-bit input for each condition, and all scores are computed in the same evaluation pipeline.

\subsection{Metrics and Alignment}
\label{sec:align}

We report PU21-PSNR, PU21-VSI, and PU21-PIQE, following the SI-HDR evaluation study~\cite{hanji2022}, together with HDR-VDP-3~\cite{mantiuk2023hdrvdp3} (v3.0.7, quality task, in JOD). ExpandDiff predictions are decoded to linear HDR first, so every method is encoded identically. We also report FID~\cite{heusel2017} on Reinhard-tone-mapped crops~\cite{reinhard2002}, denoted FID-R, following LEDiff~\cite{lediff2025}. PSNR, VSI and HDR-VDP-3 measure agreement with the reference, whereas PIQE is no-reference and FID-R compares crop statistics, so neither necessarily ranks methods by reconstruction accuracy. Crop sampling details are given in the supplementary material.

To account for global brightness differences between predictions and the reference, we fit a scalar gain $s=\sum_{k\in\Omega}p_k g_k/\sum_{k\in\Omega}p_k^2$, where $p$ is the prediction, $g$ the reference, and $\Omega$ contains RGB samples at input pixels clipped at neither end. All metrics use the aligned prediction $sp$. We additionally report full-reference metrics after SI-HDR's ground-truth-fitted 20-parameter tone-and-color correction, labeled CRF~\cite{hanji2022}. The difference between the two columns shows how much error a more flexible tone-and-color fit reduces. These are evaluation-time alignments, applied independently to every method. Reported PSNR margins are paired means and can differ slightly from subtraction of rounded table entries.

\begin{table}[!t]
\centering
\caption{Results on 181 SI-HDR references under two input conditions. PSNR, VSI, and PIQE use PU21; VDP-3 is in JOD. All scores include brightness alignment; CRF additionally applies the benchmark's tone-and-color correction. FID denotes FID-R. Best values per condition are bold and second-best values underlined.}
\label{tab:results}
\small
\setlength{\tabcolsep}{0pt}
\begin{tabular}{@{}l@{\hspace{4pt}}c@{\hspace{3.5pt}}c@{\hspace{5.75pt}}c@{\hspace{3.3pt}}c@{\hspace{1.2pt}}c@{\hspace{0.1pt}}c@{\hspace{6.05pt}}c@{\hspace{4.1pt}}c@{}}
\toprule
& \multicolumn{2}{c}{PSNR $\uparrow$} & \multicolumn{2}{c}{VSI $\uparrow$} & \multicolumn{2}{c}{VDP-3 $\uparrow$} & PIQE & FID \\
\cmidrule(lr){2-3}\cmidrule(lr){4-5}\cmidrule(lr){6-7}
Method & gain & CRF & gain & CRF & gain & CRF & $\downarrow$ & $\downarrow$ \\
\midrule
\multicolumn{9}{l}{\condnf{}} \\
\bd{ExpandDiff-B} & \bd{22.61} & \bd{30.15} & \bd{.9866} & \bd{.9897} & \phantom{0}\bd{9.06} & \phantom{0}\bd{9.46} & 28.22 & \bd{8.76} \\
\bd{ExpandDiff-P} & \underline{21.41} & \underline{28.65} & \underline{.9801} & .9867 & \phantom{0}\underline{8.77} & \phantom{0}9.35 & \bd{27.32} & 9.99 \\
\emph{LDR input} & 20.47 & 27.93 & .9756 & .9826 & \phantom{0}8.44 & \phantom{0}9.15 & 28.15 & 10.00 \\
ExpandNet & 19.17 & 28.09 & .9743 & .9840 & \phantom{0}8.37 & \phantom{0}9.12 & 28.60 & 10.65 \\
DITM & 19.06 & 27.64 & .9767 & .9857 & \phantom{0}8.52 & \phantom{0}9.00 & 30.43 & 10.70 \\
LEDiff-HL & 18.95 & 26.45 & .9722 & .9803 & \phantom{0}8.33 & \phantom{0}8.97 & 29.17 & 12.98 \\
MaskHDR & 18.77 & 28.51 & .9767 & \underline{.9874} & \phantom{0}8.59 & \phantom{0}\underline{9.37} & 29.20 & 11.92 \\
LEDiff-SH & 18.75 & 26.16 & .9708 & .9781 & \phantom{0}8.33 & \phantom{0}8.84 & \underline{27.76} & 12.19 \\
Refusion-HDR & 16.85 & 27.77 & .9555 & .9804 & \phantom{0}8.10 & \phantom{0}9.08 & 45.24 & \underline{9.58} \\
\midrule
\multicolumn{9}{l}{\condp{}} \\
\bd{ExpandDiff-P} & \bd{31.21} & \bd{33.52} & \bd{.9913} & \bd{.9917} & \phantom{0}\bd{9.62} & \phantom{0}\bd{9.68} & 39.31 & \bd{4.78} \\
MaskHDR & \underline{23.88} & \underline{28.72} & \underline{.9806} & \underline{.9884} & \phantom{0}\underline{9.28} & \phantom{0}\underline{9.65} & 38.78 & 7.88 \\
\bd{ExpandDiff-B} & 23.13 & 28.27 & .9754 & .9845 & \phantom{0}8.98 & \phantom{0}9.42 & 39.06 & 8.99 \\
ExpandNet & 22.74 & 28.34 & .9781 & .9828 & \phantom{0}8.76 & \phantom{0}9.20 & 40.17 & 6.89 \\
\emph{LDR input} & 22.23 & 27.37 & .9699 & .9754 & \phantom{0}8.49 & \phantom{0}9.00 & 40.11 & \underline{6.75} \\
LEDiff-HL & 20.99 & 25.18 & .9633 & .9771 & \phantom{0}8.29 & \phantom{0}8.90 & \underline{37.81} & 12.41 \\
DITM & 20.50 & 25.98 & .9727 & .9799 & \phantom{0}8.36 & \phantom{0}8.79 & \bd{36.83} & 11.66 \\
Refusion-HDR & 19.46 & 25.82 & .9629 & .9757 & \phantom{0}8.34 & \phantom{0}9.04 & 44.24 & 11.89 \\
LEDiff-SH & 19.14 & 25.37 & .9659 & .9733 & \phantom{0}8.45 & \phantom{0}8.80 & 37.90 & 9.73 \\
\bottomrule
\end{tabular}
\end{table}

\subsection{SI-HDR Benchmark}
\label{sec:bench}

On \condnf{} (Table~\ref{tab:results}), ExpandDiff-B reaches $22.61$\,dB gain-aligned PU21-PSNR, exceeding ExpandNet, the strongest competing model on this metric, by a paired $3.43$\,dB. ExpandDiff-P ranks second at $21.41$\,dB despite its different training degradation, with a paired advantage of $2.24$\,dB over ExpandNet. B also leads gain-aligned VSI and HDR-VDP-3, while P obtains the lowest PIQE. Together, the two variants attain the best values across all eight columns.

The advantage persists after CRF correction: ExpandDiff-B exceeds the strongest competing model by $1.64$\,dB PSNR, $0.0023$ VSI, and $0.09$ JOD. ExpandDiff-P exceeds all competitors in corrected PSNR, although MaskHDR scores slightly higher in corrected VSI and HDR-VDP-3. The smaller PSNR margin after CRF correction indicates that tone and color differences contribute to B's gain-only advantage, while its remaining lead supports an improvement beyond these correctable differences. The unprocessed input outperforms every competing model in gain-aligned PSNR, but both of our variants exceed it, and P improves on it in all eight columns. Since most input pixels remain unclipped, the strong input baseline emphasizes the need to preserve visible content while reconstructing missing information. The reported whole-image scores measure both effects. Fig.~\ref{fig:qualitative} shows two highlight-clipped examples. Additional qualitative results are in the supplementary material.

\begin{figure}[t]
\centering
\setlength{\tabcolsep}{0pt}
\renewcommand{\arraystretch}{1.0}
\newcommand{\fw}{0.199\columnwidth}
\begin{tabular}{@{}ccccc@{}}
\vspace{-0.1cm}
\includegraphics[width=\fw]{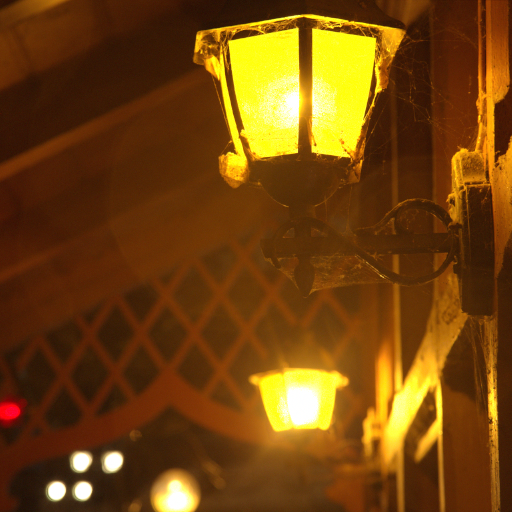} &
\includegraphics[width=\fw]{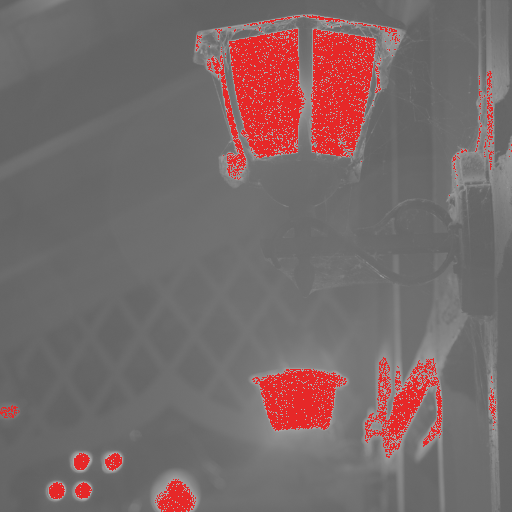} &
\includegraphics[width=\fw]{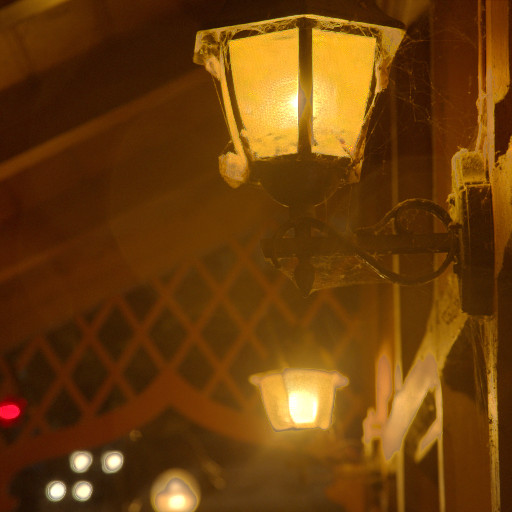} &
\includegraphics[width=\fw]{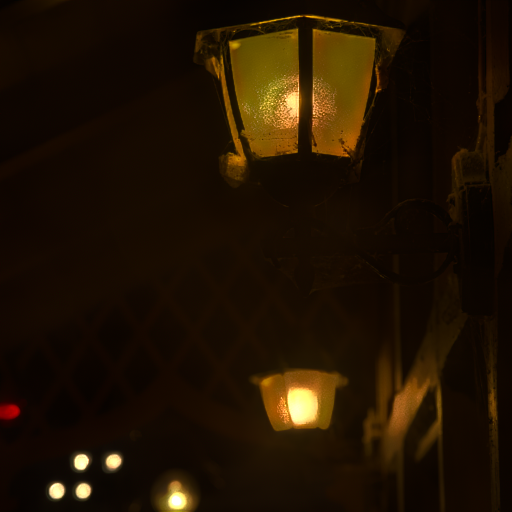} &
\includegraphics[width=\fw]{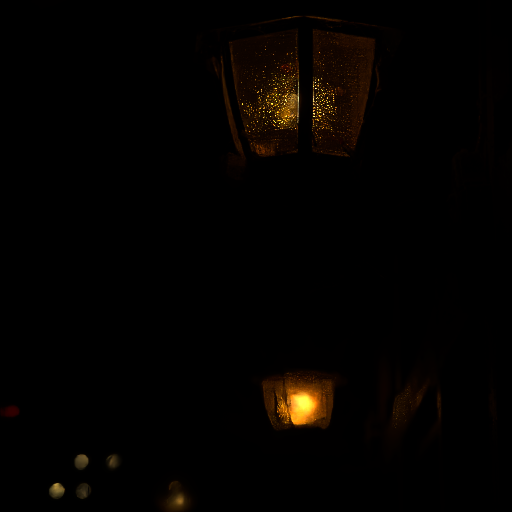} \\ 
clipped input & clip mask & LEDiff-HL & DITM & Refusion \\
\vspace{-0.1cm}
\includegraphics[width=\fw]{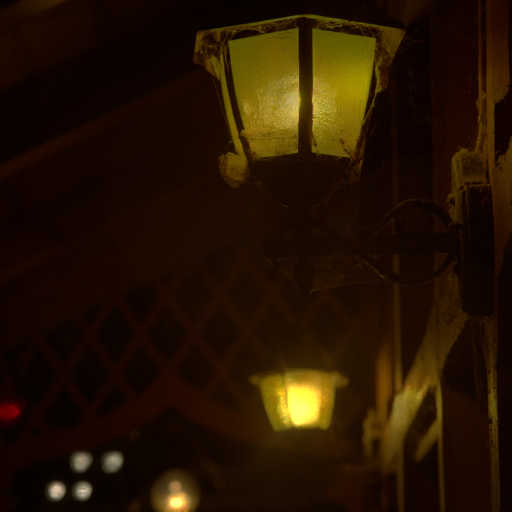} &
\includegraphics[width=\fw]{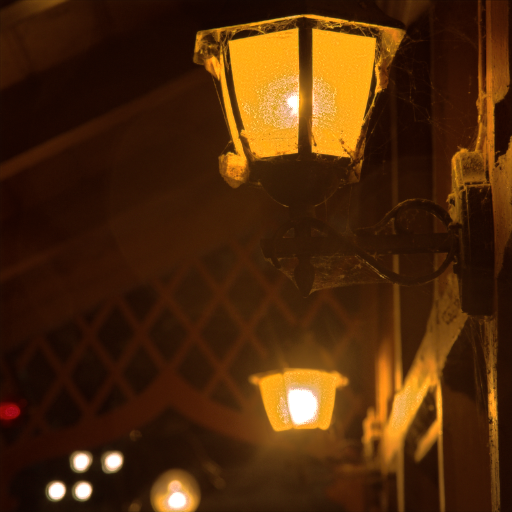} &
\includegraphics[width=\fw]{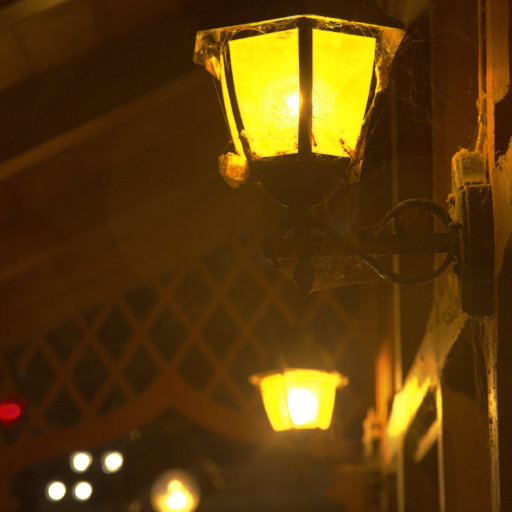} &
\includegraphics[width=\fw]{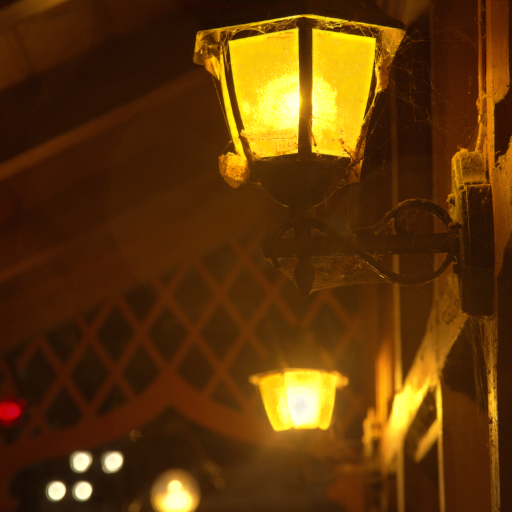} &
\includegraphics[width=\fw]{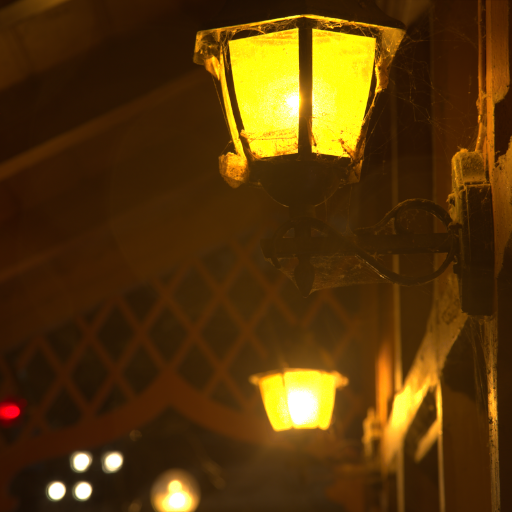} \\
MaskHDR & ExpandNet & \bd{Ours-B} & \bd{Ours-P} & ground truth \\[0.5pt]
\vspace{-0.1cm}
\includegraphics[width=\fw]{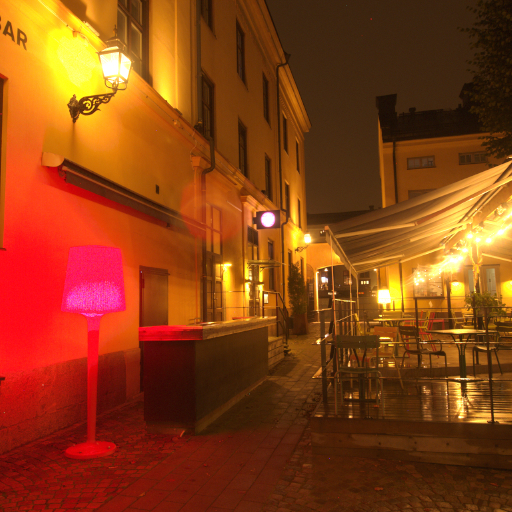} &
\includegraphics[width=\fw]{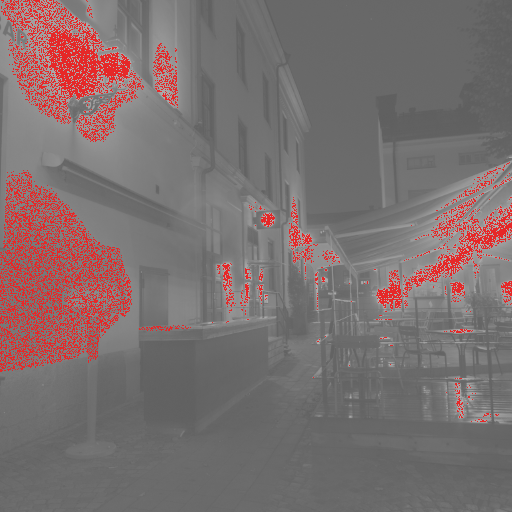} &
\includegraphics[width=\fw]{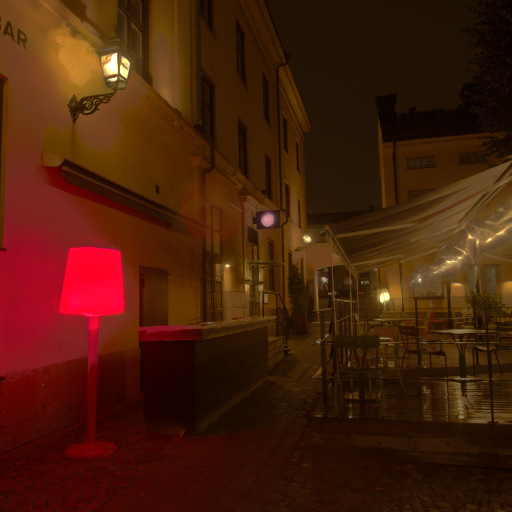} &
\includegraphics[width=\fw]{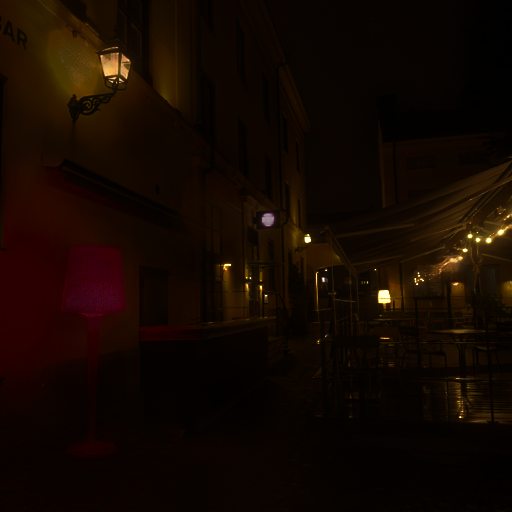} &
\includegraphics[width=\fw]{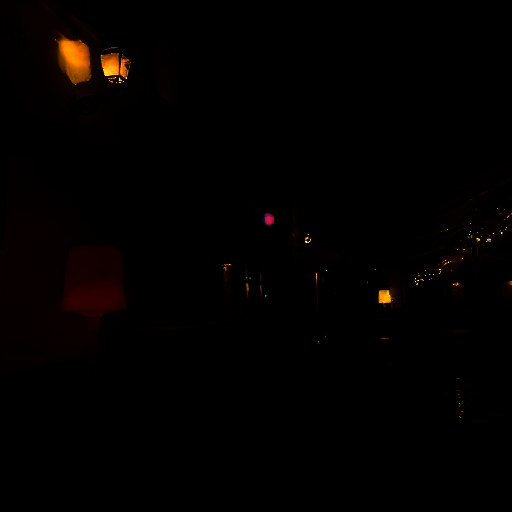} \\
clipped input & clip mask & LEDiff-HL & DITM & Refusion \\
\vspace{-0.1cm}
\includegraphics[width=\fw]{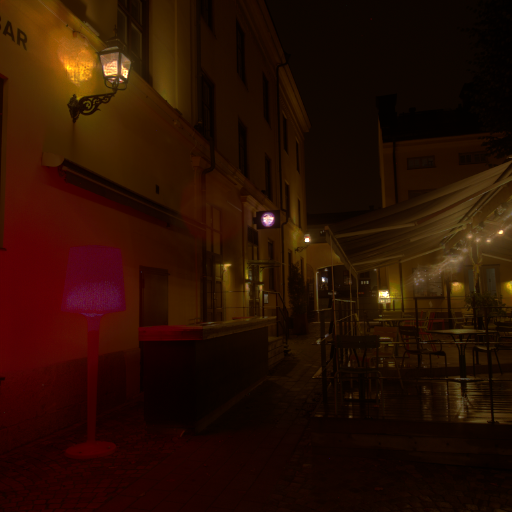} &
\includegraphics[width=\fw]{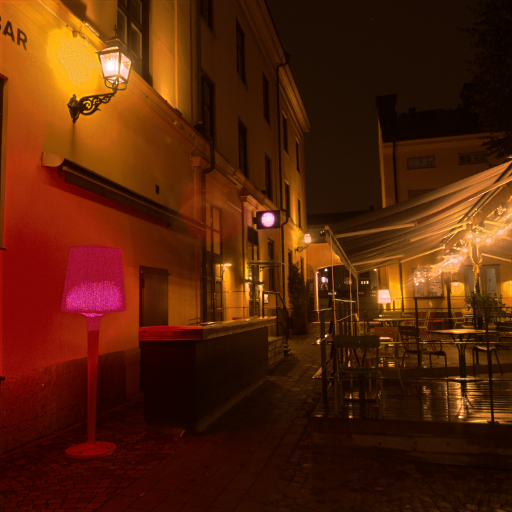} &
\includegraphics[width=\fw]{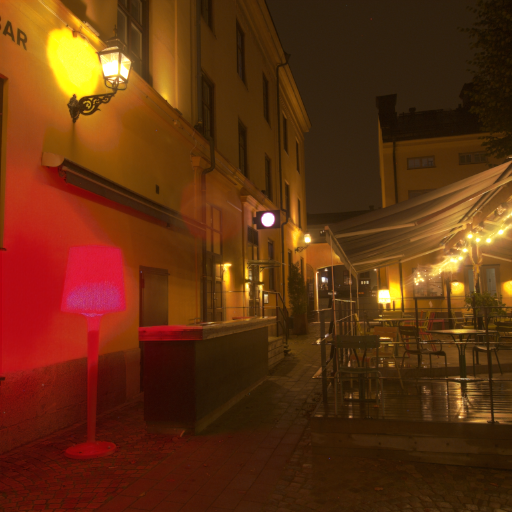} &
\includegraphics[width=\fw]{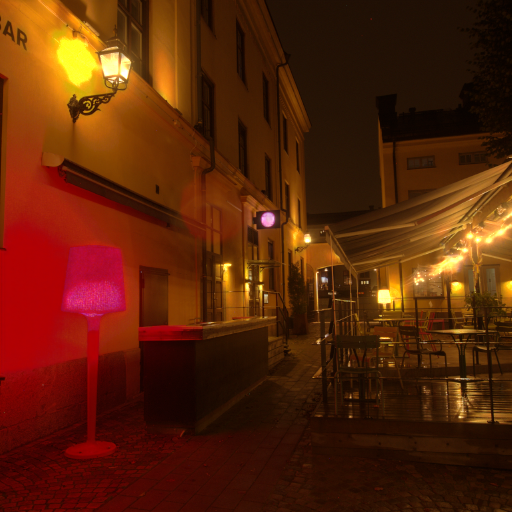} &
\includegraphics[width=\fw]{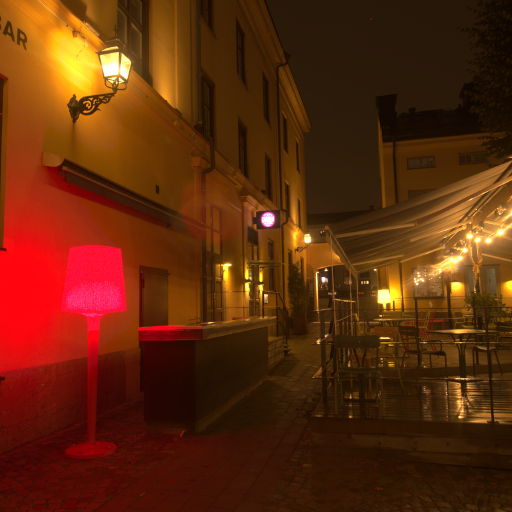} \\
MaskHDR & ExpandNet & \bd{Ours-B} & \bd{Ours-P} & ground truth
\end{tabular}
\caption{Highlight reconstruction on two \condnf{} inputs with $8.3\%$ and $6.9\%$ clipped highlights. Red marks highlight clipping. Each example occupies two rows. Reconstructions use gain-only alignment.}
\label{fig:qualitative}
\end{figure}

\subsection{Clipping Severity and Distribution Transfer}
\label{sec:twosided}
\label{sec:traindist2}

\begin{figure}[t]
\centering
\setlength{\tabcolsep}{0pt}
\renewcommand{\arraystretch}{1.0}
\newcommand{\fwc}{0.199\columnwidth}
\begin{tabular}{@{}ccccc@{}}
\vspace{-0.1cm}
\includegraphics[width=\fwc]{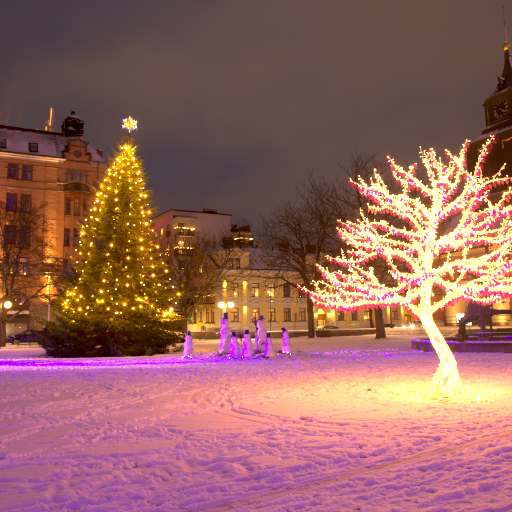} &
\includegraphics[width=\fwc]{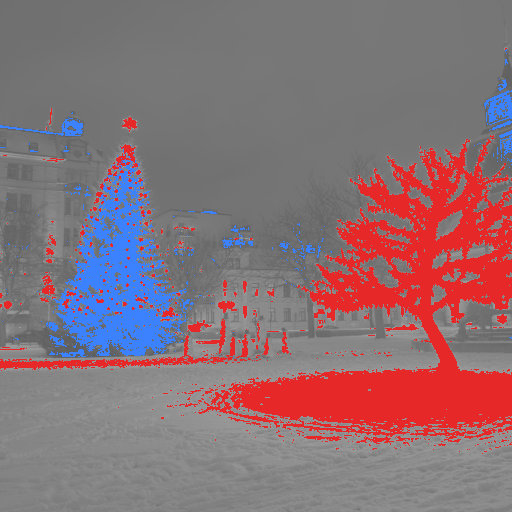} &
\includegraphics[width=\fwc]{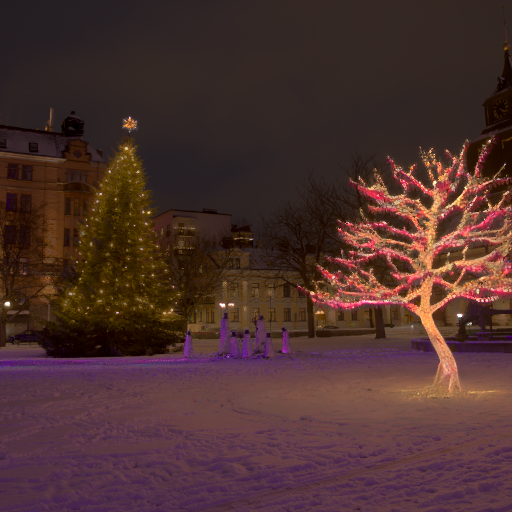} &
\includegraphics[width=\fwc]{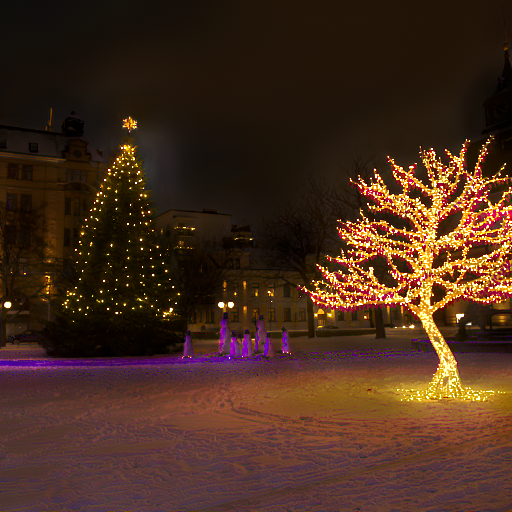} &
\includegraphics[width=\fwc]{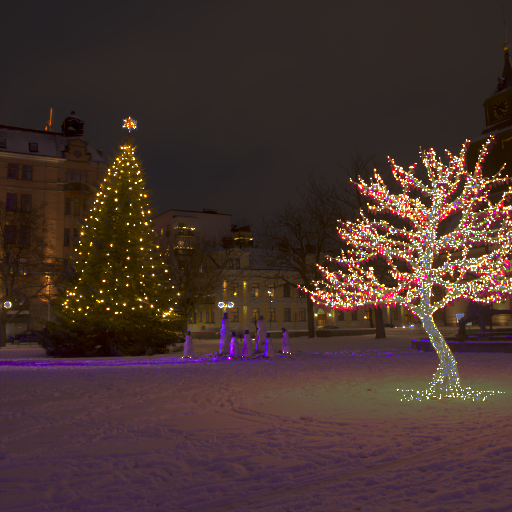} \\
clipped input & clip mask & LEDiff-HL & DITM & Refusion \\
\vspace{-0.1cm}
\includegraphics[width=\fwc]{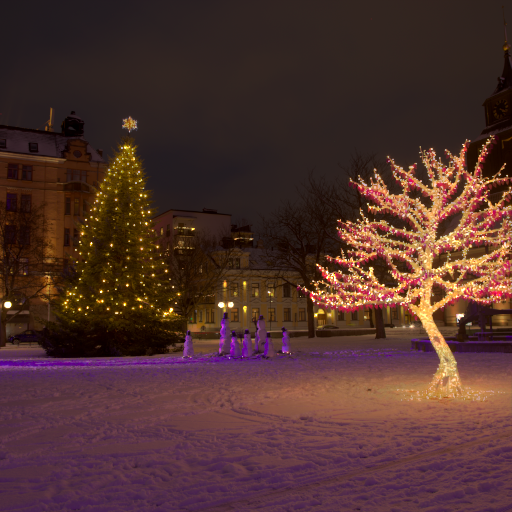} &
\includegraphics[width=\fwc]{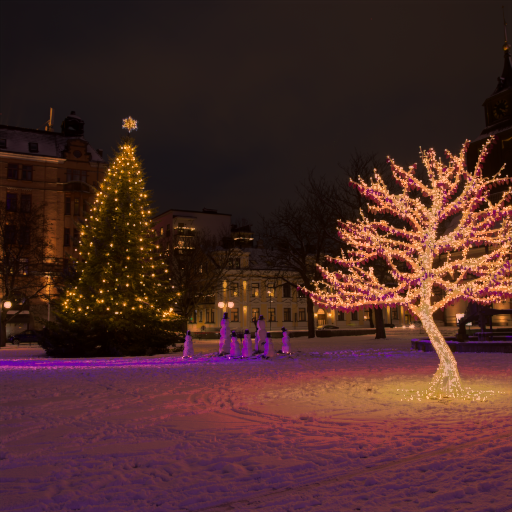} &
\includegraphics[width=\fwc]{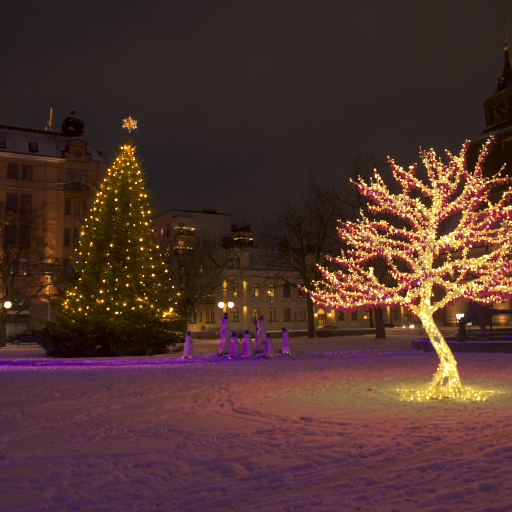} &
\includegraphics[width=\fwc]{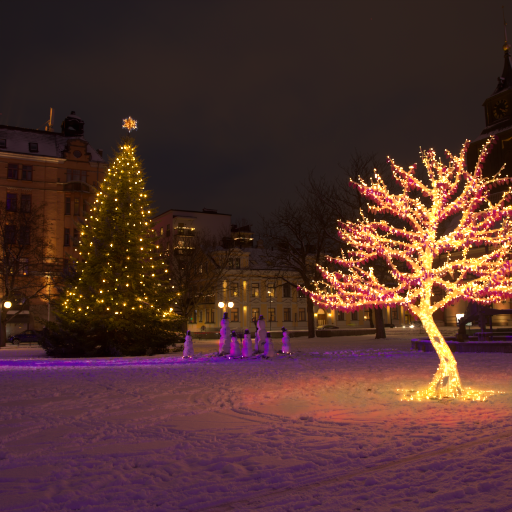} &
\includegraphics[width=\fwc]{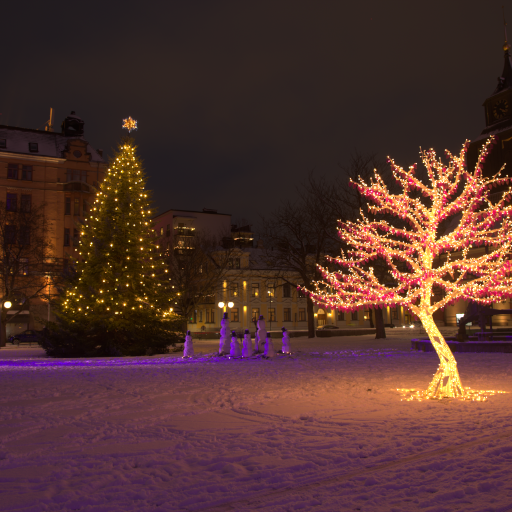} \\
MaskHDR & ExpandNet & \bd{Ours-B} & \bd{Ours-P} & ground truth \\[0.5pt]
\vspace{-0.1cm}
\includegraphics[width=\fwc]{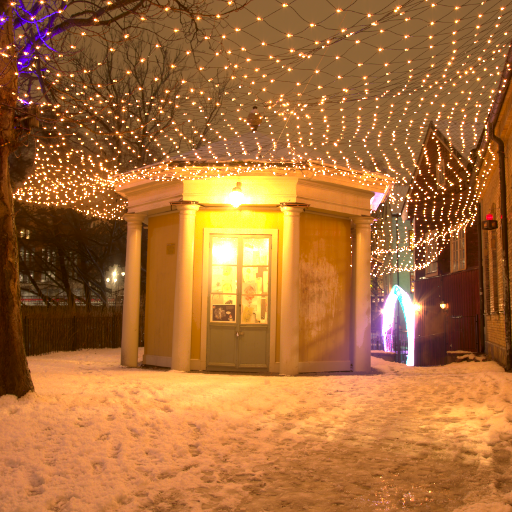} &
\includegraphics[width=\fwc]{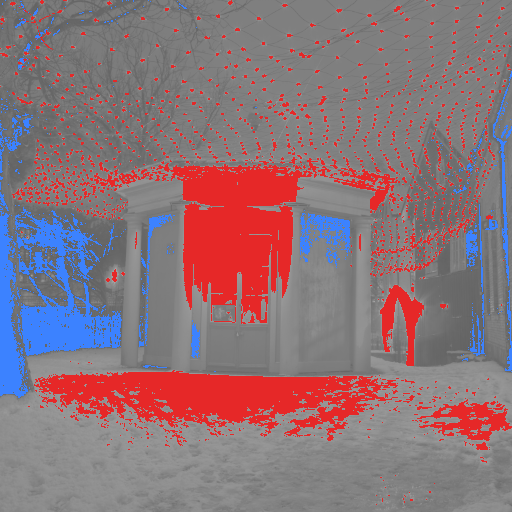} &
\includegraphics[width=\fwc]{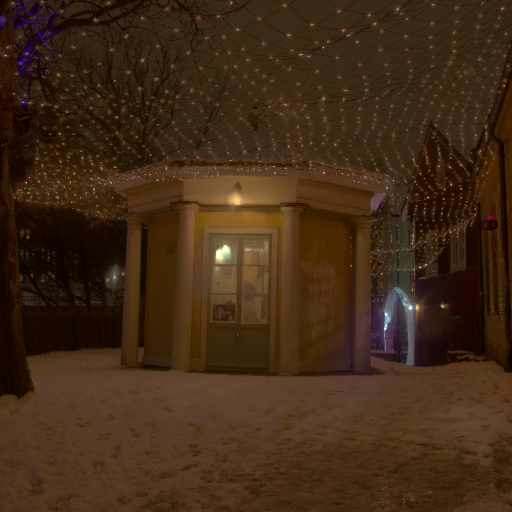} &
\includegraphics[width=\fwc]{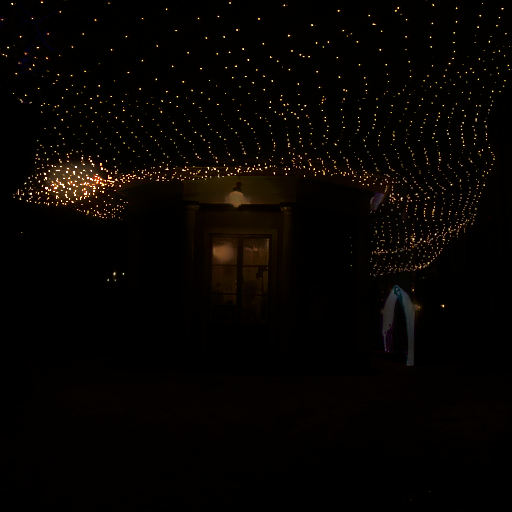} &
\includegraphics[width=\fwc]{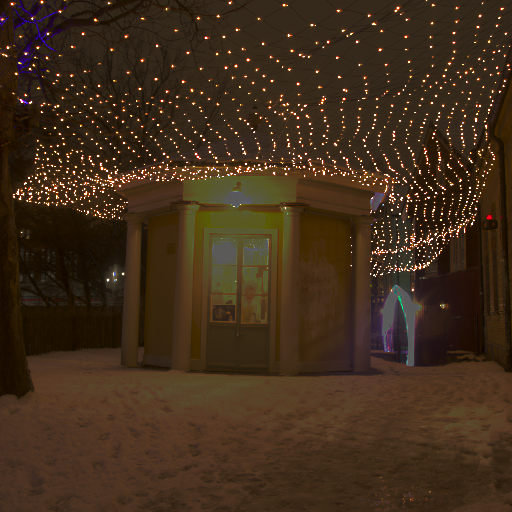} \\
clipped input & clip mask & LEDiff-HL & DITM & Refusion \\
\vspace{-0.1cm}
\includegraphics[width=\fwc]{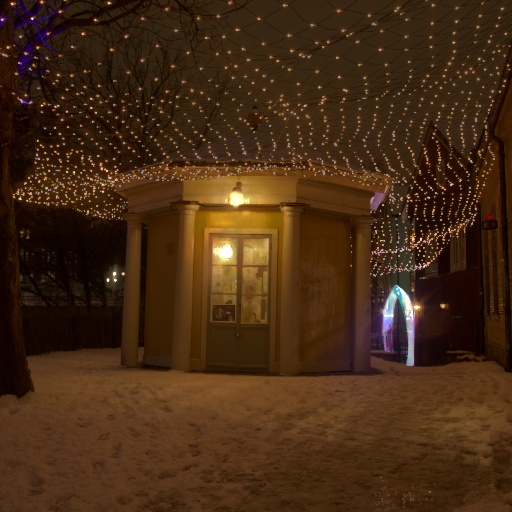} &
\includegraphics[width=\fwc]{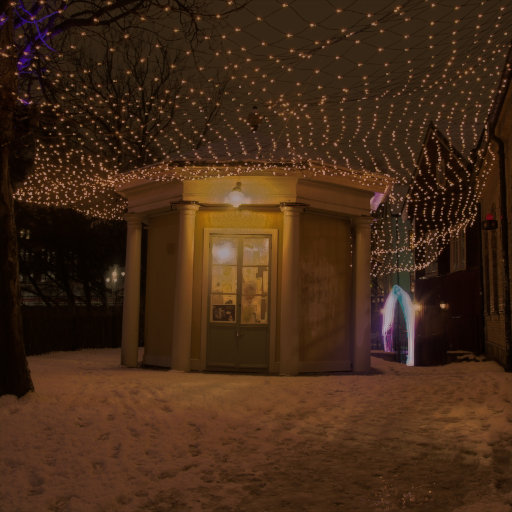} &
\includegraphics[width=\fwc]{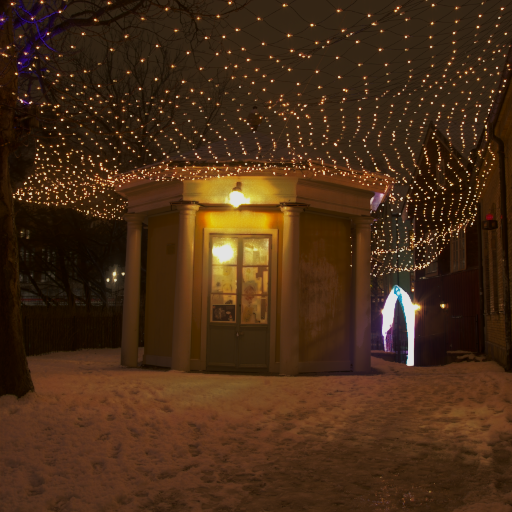} &
\includegraphics[width=\fwc]{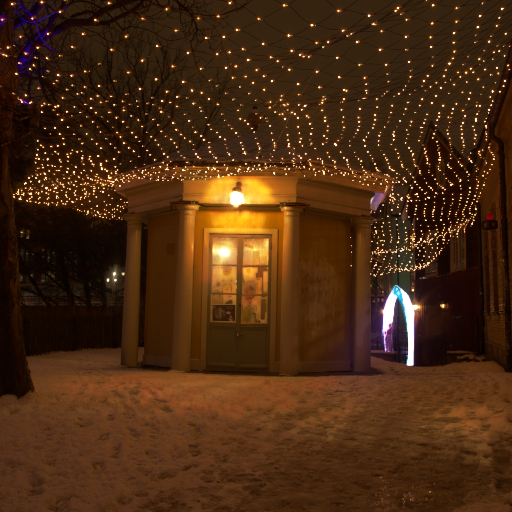} &
\includegraphics[width=\fwc]{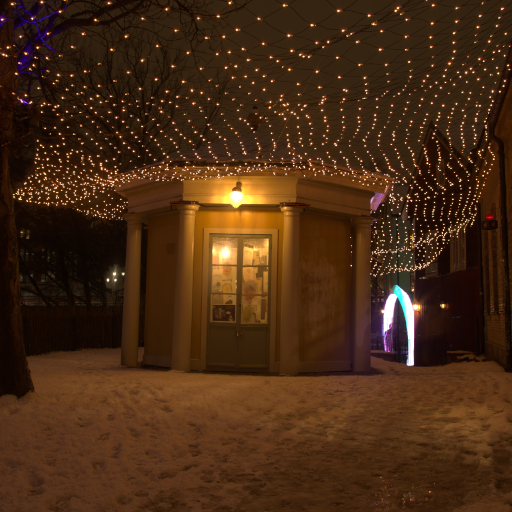} \\
MaskHDR & ExpandNet & \bd{Ours-B} & \bd{Ours-P} & ground truth
\end{tabular}
\caption{Joint shadow and highlight reconstruction on two \condp{} inputs, clipped over $20.0\%$ and $20.5\%$ of the frame. Red and blue mark clipped highlights and shadows, respectively. Each example occupies two rows. Reconstructions use gain-only alignment.}
\label{fig:twosided}
\end{figure}

On \condp{} (Table~\ref{tab:results}), ExpandDiff-P achieves $31.21$\,dB gain-aligned PSNR, a paired improvement of $7.34$\,dB over MaskHDR, and ranks first in every column except PIQE. Its HDR-VDP-3 and FID-R margins over the best competing models are $0.34$ JOD and $2.11$, respectively. It also leads all three CRF-corrected metrics. The agreement across reference-based metrics supports improved fidelity under simultaneous shadow and highlight loss. DITM obtains a lower PIQE despite substantially lower PSNR, illustrating that no-reference quality and reconstruction accuracy can favor different outputs. Fig.~\ref{fig:twosided} illustrates reconstruction on two examples when shadows and highlights are both clipped.

As reported in the supplementary material, increasing the clipping percentiles to $(10,30)$ reduces P's PSNR advantage to $4.76$\,dB, while it keeps a $0.33$ JOD advantage in HDR-VDP-3. On LEDiff's independently defined degradation, B leads the strongest competing model by $2.44$\,dB PSNR and $0.30$ JOD. The harder clipping condition tests P at the upper endpoints of its training ranges, while LEDiff's degradation tests a different input synthesis pipeline. The retained margins show that the gains are not confined to the midpoint clipping condition.

The P/B ranking reverses across input distributions: B leads P on seven metrics at \condnf{}, and P leads B on seven at \condp{}. The comparison at their reported checkpoints also differs in training duration and learning rate. A matched-budget check at 75k steps gives P $21.38$\,dB PSNR on \condnf{}, still below B's $22.61$\,dB. This supports matching the training degradation to the benchmark, while P's second place demonstrates transfer from the DCS distribution.

\subsection{Target Encoding and Output Head}
\label{sec:ablation}

We train all four combinations of linear or PU21 targets and bounded or unbounded heads for 150k steps, holding the remaining P configuration fixed. Table~\ref{tab:ablation} shows that removing PU21 reduces gain-aligned PSNR by $2.00$\,dB, removing $\tanh$ by $1.01$\,dB, and removing both by $4.37$\,dB. The full model also leads the CRF-corrected fidelity metrics. Each component improves PSNR with either setting of the other component, supporting both choices within this architecture. Their combined benefit is consistent with the prediction-range argument in Sec.~\ref{sec:pu21}, although the ablation does not separate the effects of perceptual error weighting and output sensitivity. This benefit is metric-dependent: the unbounded PU21 model has a slightly lower FID-R, and the unbounded linear model has the lowest PIQE.

\begin{table}[t]
\centering
\caption{PU21 encoding and the bounded head on \condp{}. All variants use the same data and 150k training steps. Metric definitions and emphasis follow Table~\ref{tab:results}.}
\label{tab:ablation}

\small
\setlength{\tabcolsep}{0pt}
\begin{tabular}{@{}l@{\hspace{4pt}}c@{\hspace{3.5pt}}c@{\hspace{5.75pt}}c@{\hspace{3.3pt}}c@{\hspace{1.2pt}}c@{\hspace{0.1pt}}c@{\hspace{6.05pt}}c@{\hspace{4.1pt}}c@{}}
\toprule
& \multicolumn{2}{c}{PSNR $\uparrow$} & \multicolumn{2}{c}{VSI $\uparrow$} & \multicolumn{2}{c}{VDP-3 $\uparrow$} & PIQE & FID \\
\cmidrule(lr){2-3}\cmidrule(lr){4-5}\cmidrule(lr){6-7}
Method & gain & CRF & gain & CRF & gain & CRF & $\downarrow$ & $\downarrow$ \\
\midrule
\bd{ExpandDiff-P} & \bd{31.21} & \bd{33.52} & \bd{.9913} & \bd{.9917} & \phantom{0}\bd{9.62} & \phantom{0}\bd{9.68} & \underline{39.31} & \phantom{0}\underline{4.78} \\
PU21, no $\tanh$ & \underline{30.20} & \underline{33.08} & \underline{.9905} & \underline{.9910} & \phantom{0}\underline{9.59} & \phantom{0}\underline{9.67} & 39.46 & \phantom{0}\bd{4.77} \\
linear, $\tanh$ & 29.21 & 31.38 & .9890 & .9897 & \phantom{0}9.51 & \phantom{0}9.60 & 40.04 & \phantom{0}6.03 \\
linear, no $\tanh$ & 26.84 & 30.34 & .9861 & .9880 & \phantom{0}9.46 & \phantom{0}9.61 & \bd{38.16} & \phantom{0}5.56 \\
\bottomrule
\end{tabular}
\end{table}

\noindent \textbf{Inference cost.} ExpandDiff takes $3.19$\,s per $512\times512$ image on an NVIDIA A100 80\,GB with 24 DDIM steps. The fully convolutional network accepts other image sizes after reflection padding to multiples of 64, subject to available memory. Increasing sampling to 48 or 100 steps gives no clear PSNR improvement as reported in the supplementary material.

\section{Conclusion}

We introduced ExpandDiff, which couples Dynamic Clipping Synthesis, spatial LDR guidance, and perceptually encoded diffusion for joint shadow and highlight reconstruction. Its results across SI-HDR and additional clipping conditions show that a broadly sampled training distribution can transfer beyond its nominal degradation, while a benchmark-oriented distribution can further improve matched performance. The target ablation supports using PU21 with a bounded head for reconstruction fidelity. Like other methods, ExpandDiff cannot know the scene's absolute brightness, and its reconstructions weaken over large contiguous clipped regions. Further limitations are discussed in the supplementary material.

\bibliographystyle{IEEEbib}
\bibliography{refs}

\end{document}